\documentclass[10pt,twocolumn,letterpaper]{article}

 \usepackage{cvpr}              

\usepackage{ragged2e} 
\usepackage{booktabs,makecell, multirow, tabularx}
\usepackage{pifont}											
\usepackage{colortbl}  
\usepackage{array}   
\usepackage{subcaption} 
\usepackage{float}											

\usepackage[dvipsnames]{xcolor}

\definecolor{cvprblue}{rgb}{0.21,0.49,0.74}
\usepackage[pagebackref,breaklinks,colorlinks,citecolor=cvprblue]{hyperref}

\def\paperID{2076} 
\def\confName{CVPR}
\def\confYear{2024}

\title{Traffic Sign Interpretation in Real Road Scene}

\author{Chuang Yang$\dagger$\\
	NWPU, SCS, iOPEN\\
	{\tt\small omtcyang@gmail.com}
	\and
	Kai Zhuang$\dagger$\\
	NWPU, iOPEN\\
	\and
	Mulin Chen\\
	NWPU, iOPEN\\
	\and
	Haozhao Ma\\
	NWPU, SS, iOPEN\\
	\and
	Xu Han\\
	NWPU, SCS, iOPEN\\
	\and
	Tao Han\\
	Shanghai AI Lab\\
	\and
	Changxing Guo\\
	NWPU, SCS, iOPEN\\
	\and
	Han Han\\
	NWPU, iOPEN\\
	\and
	Bingxuan Zhao\\
	NWPU, iOPEN\\
	\and
	Qi Wang*\\
	NWPU, iOPEN\\
}

\begin{document}
\maketitle
\begin{abstract}
Most existing traffic sign-related works are dedicated to detecting and recognizing part of traffic signs individually, which fails to analyze the global semantic logic among signs and may convey inaccurate traffic instruction. Following the above issues, we propose a traffic sign interpretation (TSI) task, which aims to interpret global semantic interrelated traffic signs (e.g.,~driving instruction-related texts, symbols, and guide panels) into a natural language for providing accurate instruction support to autonomous or assistant driving. Meanwhile, we design a multi-task learning architecture for TSI, which is responsible for detecting and recognizing various traffic signs and interpreting them into a natural language like a human. Furthermore, the absence of a public TSI available dataset prompts us to build a traffic sign interpretation dataset, namely TSI-CN. The dataset consists of real road scene images, which are captured from the highway and the urban way in China from a driver's perspective. It contains rich location labels of texts, symbols, and guide panels, and the corresponding natural language description labels. Experiments on TSI-CN demonstrate that the TSI task is achievable and the TSI architecture can interpret traffic signs from scenes successfully even if there is a complex semantic logic among signs. The TSI-CN dataset and the source code of the TSI architecture will be publicly available after the revision process.
\end{abstract}    
\begin{figure}[t]
	\centering
	\includegraphics[width=0.99\linewidth]{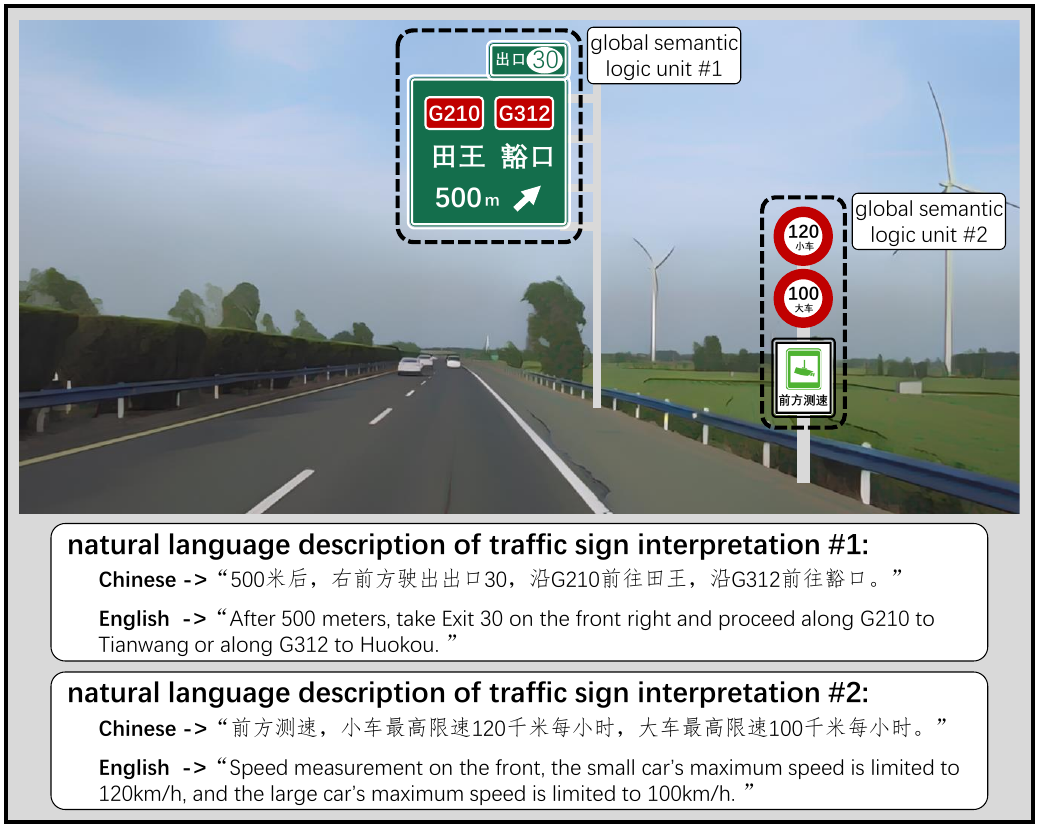}
	\vspace{-.1in}
	\caption{The illustration of TSI task. Given an RGB real road scene image, TSI detects and recognizes traffic texts, symbols, and guide panels at first. It then organizes the global semantic logic among these components to generate natural language descriptions with accurate traffic instruction information.}
	\label{fig:V1}
	\vspace{-.15in}
\end{figure}

\section{Introduction}
\label{sec:intro}
Intelligent transportation has progressed in recent years with the rapid development of deep learning technology. As a fundamental task in the field of intelligent transportation, traffic sign understanding has become a hot topic and attracted more attention. 

Traffic signs consist of driving instruction-related symbols, texts, and guide panels. Initially, researchers mainly focused on the recognition of traffic symbols individually~\cite{DBLP:conf/cvpr/ZhuLZHLH16,DBLP:journals/tits/MinLHHWW22}, which makes it hard to understand traffic sign information accurately and even results in inappropriate driving behaviors. For example, the detection and classification of the speed limit symbol, one of the most common traffic signs, has already played an important role in current autonomous or assistant driving systems. However, the symbol often appears along with other signs to convey road traffic information as a whole (e.g., the combination of speed limit symbol and vehicle type information). Therefore, driving according to the information obtained from speed limit signs alone may cause traffic jams or even accidents. Though some works propose to recognize traffic texts~\cite{rong2016recognizing,DBLP:journals/tits/HouZLYWWY21}, considering all signs together for understanding road traffic information is still under researched. Recently, some researchers have tried to predict the relationships among traffic directional marks and texts~\cite{guo2021learning,guo2023visual}. Though they have achieved progress compared with previous related works, putting directional marks and texts together simply without the analysis of global semantic logic among signs makes it still hard to interpret useful and accurate traffic instruction information from real road scenes.

Considering the above problems, the task of traffic sign interpretation (TSI) is proposed in this work for providing useful and accurate traffic instruction information support to autonomous and assistant driving systems. Different from~\cite{guo2021learning,guo2023visual}, the TSI task is defined as interpreting interrelated signs into natural languages like a human. To achieve the interpretation task, we propose a TSI architecture. It locates and recognizes traffic symbols, texts, and guide panels from real traffic scenes at first. The architecture then analyzes the global semantic logic according to the location and recognition results of signs. In the end, natural language that can convey traffic instruction information is organized. 

Furthermore, the absence of a public TSI available dataset prompts us to construct a related dataset for promoting the progress of the sign interpretation community. We build a traffic sign interpretation dataset from China, namely TSI-CN, to fulfill the research and evaluation in this field. Compared with the previous \textbf{traffic sign recognition dataset}~\cite{stallkamp2012man,yang2015towards,tabernik2019deep} and \textbf{traffic sign understanding dataset}~\cite{guo2021learning,guo2023visual}, TSI-CN enjoys the following two main advantages over them: 1) Containing the location and recognition labels of traffic symbols, texts, and guide panels simultaneously; 2) Natural language descriptions based on the analysis of global semantic logic among signs. The contributions of this work are summarized as follows:

\begin{itemize}
	\item[1.] The traffic sign interpretation (TSI) task is proposed to interpret interrelated signs into natural languages based on the analysis of global semantic logic among them. It helps understand and convey accurate instruction information from real road scenes like a human.

	\item[2.] A TSI task corresponding dataset (TSI-CN) is built. It is the first traffic sign dataset from real road scenes of China, which is equipped with location and recognition labels of various signs and natural language description labels of logic-interrelated signs simultaneously.
	
	\item[3.] A TSI architecture is designed, which consists of the three sub-structures of detection, recognition, and interpretation. Experimental results on the created TSI-CN dataset demonstrate that the TSI task is achievable and the effectiveness of TSI architecture. 
	
	\item[4.] The TSI task and the corresponding dataset will promote the progress of the traffic sign recognition community and provide support for the development of autonomous and assistant driving systems.
\end{itemize}
\section{Related Works}
\label{sec:rela}
The field of traffic sign recognition has attracted more attention recently with the rapid development of computer vision-based autonomous or assistant driving systems. Existing related works will be introduced from the two aspects of the method and dataset in this section. 
\subsection{Traffic Sign-Related Method}
Initially, researchers focused on traffic symbol recognition individually. Saturnino~\etal~\cite{maldonado2007road} classified symbols into different categories based on support vector machines (SVMs). Lu~\etal~\cite{lu2012sparse} introduced local manifold structures into sign classification via graph embedding algorithm. With the development of deep learning, many Convolutional Neural Networks (CNNs)-based traffic symbol recognition methods~\cite{cirecsan2011committee,sermanet2011traffic,jin2014traffic,li2017perceptual} are proposed. Dan~\etal~\cite{cirecsan2011committee} and Pierre~\etal~\cite{sermanet2011traffic} proposed to detect and classify symbols with an end-to-end CNN framework, which ran faster and recognized more accurately than previous traditional methods. Jin~\etal~\cite{jin2014traffic} designed a hinge loss to encourage the model to focus on hard samples. Li~\etal~\cite{li2017perceptual} utilized a Generative Adversarial Network (GAN) to represent small symbols to super-resolved ones to avoid challenges from low-resolution images.
Except for the traffic symbol recognition, Rong~\etal~\cite{rong2016recognizing} combined CNNs and Recurrent Neural Networks (RNNs) to recognize traffic texts inspired by the scene text detection and recognition technique~\cite{epshtein2010detecting,ye2014text,jaderberg2016reading}. To better understand traffic signs, Guo~\etal~\cite{guo2021learning} recognized traffic symbols and texts simultaneously and combined recognition results simply. However, interpreting accurate instruction information according to the global semantic logic among signs from real road scenes is still under researched.

\subsection{Traffic Sign-Related Dataset}
To fulfill the research and evaluation of the traffic sign recognition field, some task-related datasets~\cite{stallkamp2012man,yang2015towards,DBLP:conf/cvpr/ZhuLZHLH16,tabernik2019deep,guo2021learning} are constructed. Johannes~\etal~\cite{stallkamp2012man} collected more than 50,000 images in Germany. The dataset contained 43 different traffic symbols with rectangle box labels. Yang~\etal~\cite{yang2015towards} created a China Traffic Sign
Dataset (CTSD) with sparse symbol distribution. The images of different sizes, low-light scenes, and low resolution in CTSD encouraged researchers to focus on complex background analysis. Zhu~\etal~\cite{DBLP:conf/cvpr/ZhuLZHLH16} constructed a large-scale traffic symbol dataset, namely Tsinghua-Tencent 100K, that was sampled from Tencent Street View panoramas. Considering the deficiency of the most normal symbols, Domen~\etal~\cite{tabernik2019deep} built a traffic-sign dataset with 200 symbol categories. Considering the recognition of symbols or texts individually could not convey integrity traffic instruction information, Guo~\etal~\cite{guo2021learning,guo2023visual} cropped traffic guide panels patches from images to construct a traffic sign understanding dataset. It is equipped with labeled arrows, texts, and the relationships among them. However, a traffic sign dataset that is captured from real road scenes while containing labeled symbols, texts, guide panels, and natural language descriptions of traffic instruction information based on the analysis of global semantic logic among signs simultaneously is needed. It will promote the progress of the traffic sign recognition community and provide support for the development of autonomous and assistant driving systems.

\section{TSI Task Description}
Different from previous works~\cite{guo2021learning} that recognize traffic directional marks and predict the relationship among them simply, the TSI task aims to convey accurate traffic instruction information via natural language $l$ by finding out all signs $q=\{s,t,p\}$ from real road scenes and organizing the global semantic logic $h$ among them. The whole interpretation process of traffic signs can be described as:
\begin{eqnarray}
h=\mathcal{H}(q),\\
l=\mathcal{T}(h,q),
\end{eqnarray} 
where $q$ is a collection with detected and recognized symbols ($s$), texts ($t$), and guide panels ($p$). $h$ is the hidden state of the global semantic logic of $q$ that generated from some nonlinear functions $\mathcal{H}(\cdot)$. $\mathcal{T}(\cdot)$ and $l$ are the interpretation mapping function and output natural language strings $l$ with traffic instruction information. The same as $\mathcal{H}(\cdot)$, $\mathcal{T}(\cdot)$ consists of nonlinear functions. $l$ can be written as:
\begin{eqnarray}
l=(c_0,c_1, ...,c_{n-1},c_n),
\end{eqnarray} 
where $c_n$ is the $n$-th character of the string $l$.   
\begin{table*}[]
	\renewcommand{\arraystretch}{.8}
	\setlength{\tabcolsep}{1mm}
	\centering
	\footnotesize   
	\caption{The essential differences between the existing sign datasets and TSI-CN. `GSL' and `NL' denote `global semantic logic' and `natural language', respectively. `RRS' denotes real road scenes. $\mathrm{symbol}_a$ and $\mathrm{symbol}_o$ are arrow symbols and the others, respectively.}
	\vspace{-.1in}
	\begin{tabular}{l|c|cccc|cccc|cc} 
		\Xhline{1.5pt}
		\multirow{2}{*}{dataset}                                      & \multirow{2}{*}{RRS}       & \multicolumn{4}{c|}{detection content}                                                                                                                                           & \multicolumn{4}{c|}{recognition content}                                                                                                                                           & \multicolumn{2}{c}{interpretation content}                                                        \\ \cline{3-12} 
		&                            & \multicolumn{1}{c}{$\mathrm{symbol}_a$}                    & \multicolumn{1}{c}{$\mathrm{symbol}_o$}              & \multicolumn{1}{c}{text}                       & panel                      & \multicolumn{1}{c}{$\mathrm{symbol}_a$}                    & \multicolumn{1}{c}{$\mathrm{symbol}_o$}              & \multicolumn{1}{c}{text}                       & panel                      & \multicolumn{1}{c}{GSL analysis}               & NL description             \\ \hline  
		GTSRB~\cite{stallkamp2012man}           &                            & \multicolumn{1}{c}{}                           & \multicolumn{1}{c}{\ding{52}} & \multicolumn{1}{c}{}                           &                            & \multicolumn{1}{c}{}                           & \multicolumn{1}{c}{\ding{52}} & \multicolumn{1}{c}{}                           &                            & \multicolumn{1}{c}{}                           &                            \\ 
		CTSD~\cite{yang2015towards}             & \ding{52} & \multicolumn{1}{c}{}                           & \multicolumn{1}{c}{\ding{52}} & \multicolumn{1}{c}{}                           &                            & \multicolumn{1}{c}{}                           & \multicolumn{1}{c}{\ding{52}} & \multicolumn{1}{c}{}                           &                            & \multicolumn{1}{c}{}                           &                            \\ 
		TT100K~\cite{DBLP:conf/cvpr/ZhuLZHLH16} & \ding{52} & \multicolumn{1}{c}{}                           & \multicolumn{1}{c}{\ding{52}} & \multicolumn{1}{c}{}                           &                            & \multicolumn{1}{c}{}                           & \multicolumn{1}{c}{\ding{52}} & \multicolumn{1}{c}{}                           &                            & \multicolumn{1}{c}{}                           &                            \\ 
		DFG~\cite{tabernik2019deep}             & \ding{52} & \multicolumn{1}{c}{}                           & \multicolumn{1}{c}{\ding{52}} & \multicolumn{1}{c}{}                           &                            & \multicolumn{1}{c}{}                           & \multicolumn{1}{c}{\ding{52}} & \multicolumn{1}{c}{}                           &                            & \multicolumn{1}{c}{}                           &                            \\ 
		CTSU~\cite{guo2021learning}             &                            & \multicolumn{1}{c}{\ding{52}} & \multicolumn{1}{c}{\ding{52}} & \multicolumn{1}{c}{\ding{52}} &                            & \multicolumn{1}{c}{\ding{52}} & \multicolumn{1}{c}{}                           & \multicolumn{1}{c}{\ding{52}} &                            & \multicolumn{1}{c}{}                           &                            \\ 
		RS10K~\cite{guo2023visual}              & \ding{52} & \multicolumn{1}{c}{\ding{52}} & \multicolumn{1}{c}{\ding{52}} & \multicolumn{1}{c}{\ding{52}} & \ding{52} & \multicolumn{1}{c}{\ding{52}} & \multicolumn{1}{c}{}                           & \multicolumn{1}{c}{\ding{52}} &                            & \multicolumn{1}{c}{}                           &                            \\ \hline
		\multicolumn{1}{c|}{TSI-CN (\textbf{Ours})} & \ding{52} & \multicolumn{1}{c}{\ding{52}} & \multicolumn{1}{c}{\ding{52}} & \multicolumn{1}{c}{\ding{52}} & \ding{52} & \multicolumn{1}{c}{\ding{52}} & \multicolumn{1}{c}{\ding{52}} & \multicolumn{1}{c}{\ding{52}} & \ding{52} & \multicolumn{1}{c}{\ding{52}} & \ding{52} \\ 
		\Xhline{1.5pt}
	\end{tabular}
\vspace{-.1in}
\label{tab:T1}
\end{table*}
\begin{figure*}[t]
	\centering
	\includegraphics[width=0.8\linewidth]{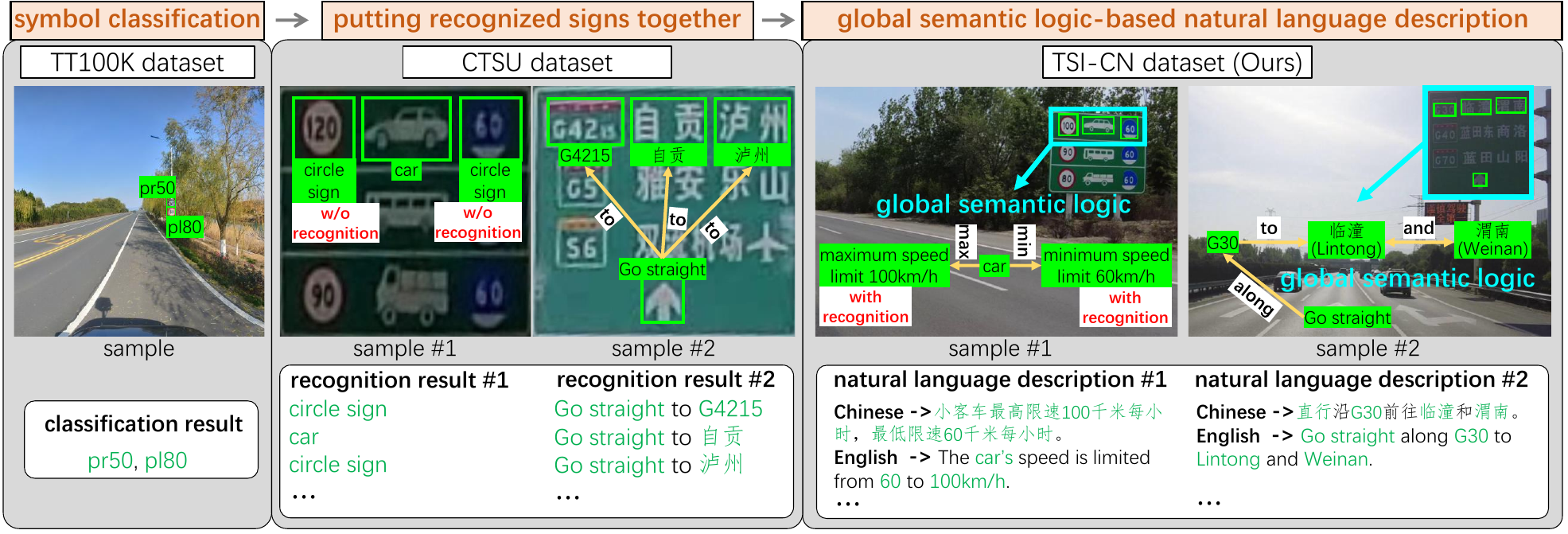}
	\vspace{-.1in}
	\caption{The illustration of the essential differences (in Table~\ref{tab:T1}) between the existing representative datasets and the TSI-CN dataset.}
	\label{fig:V2}
	\vspace{-.1in}
\end{figure*}

\section{TSI-CN Dataset}
\label{TSI-CN Dataset}
To fulfill the research and evaluation of TSI task, the TSI-CN dataset is constructed. This section describes the dataset from four perspectives: data collection, data annotation, data analysis, and data split and evaluation metric.

\subsection{Data Collection}
Our data are collected from self-shooting in the driver's perspective through the DJI OSMO Pocket2 camera. The images in the TSI-CN Dataset are captured from traffic record videos related to some popular Chinese cities, including Xi'an, Xianyang, Baoji, Zhengzhou, and Zhoukou, containing some typical road scenes, such as highways, urban expressways, urban streets, and rural roads. In the capturing process, we sample a key frame of a picture every 5 seconds from the recorded videos. Considering the traffic jam will lead to duplicate frames within a long time, the above-sampled frames per 5 seconds will be re-sampled to construct the TSI-CN dataset.

\subsection{Data Annotation}
For providing supervision labels to detection, recognition, and interpretation tasks (as shown in Figure.~\ref{fig:V2}), we annotate the collected images using the Labelme platform to generate the following three kinds of annotations: 1. Detection annotation; 2. Recognition annotation; 3. Natural language description annotation. The details of the annotation process and criterion will be illustrated following.

\textbf{Detection and recognition annotation.} Both the generation of detection and recognition labels are done together. Concretely, boxes with four corner points are drawn first to label sign (including symbol, text, and guide panel) locations to serve as the training supervision information for the detection task. Then, the recognition annotation can be labeled via the box name. Here, we design different naming methods for the symbol, text, and guide panel for a convenient labeling process. 

For the \textbf{symbol}, we divide them into two types, including $\mathrm{symbol}_a$ (\textit{\textbf{a}}rrow) and $\mathrm{symbol}_o$ (\textit{\textbf{w}}arning, \textit{\textbf{i}}nstruction, and \textit{\textbf{p}}rohibit), and naming symbol boxes via the combination string of the symbol type and serial number (e.g., `w10' and `a2'), where `w' and `a' represent the symbol type and the number refers to which particular category it is in a type.

For the \textbf{text}, the recognition annotation is the corresponding text string, which includes Arabic numerals, English, Chinese, and other special characters. Particularly, considering some strokes of texts stick to each other making it difficult to distinguish them, they are labeled as `\#\#\#' and ignored in both the training and inference processes by following the way of previous OCR datasets~\cite{ch2017total,yao2012detecting,karatzas2015icdar}.

For the \textbf{guide panel}, it is divided into seven categories according to the content and basic visual features. They are the prohibit, warning, normal road instruction, highway instruction, scenic area instruction, notice, and dynamic prompt panels, which are labeled with `1'$\sim $`7' respectively. Meanwhile, considering there are lots of noises (such as architectural graffiti, billboards, etc) that enjoy highly similar visual features with panels, distinguishing them according to traffic-related symbols and texts is an effective and essential way. Therefore, a guide panel will be ignored if all symbols and texts within it are not clearly visible.

\textbf{Natural language annotation.}
We follow the Chinese design criteria of road traffic signs to organize the global semantic logic among signs at first. Then, we put the signs that belong to the same semantic logic unit together and describe the traffic instruction information based on the global semantic logic via natural language.

\subsection{Data Analysis}
TSI-CN dataset consists of 2,682 images of 2160$\times$3840 pixels, with 2,910 natural language description sentence annotations. We divide symbols and guide panels into 7 and 155 categories respectively. Meanwhile, TSI-CN provides 144,130 text characters to encourage the text recognition task. Compared with the existing sign datasets, it is the largest from the perspective of image size and annotation number. Existing datasets can be roughly divided into sign recognition and understanding datasets. We specifically compare our TSI-CN dataset with two representative (TT100K and CTSU) datasets to show the essential differences between them in Figure.~\ref{fig:V2} (RS10K has not been publicly available until submission):

1. \textbf{Detection and recognition content.} The proposed TSI-CN dataset provides the annotation of symbols, texts, and guide panels for the sign detection and recognition task simultaneously, which supports to analysis of traffic instruction information integrally. 

2. \textbf{Interpretation content.} Different from all previous datasets, TSI-CN encourages to organization of the signs and the global semantic logic among them to express traffic instruction information via natural language, which ensures the accuracy and integrity of the interpretation of traffic instruction information.

3. \textbf{Image resolution.} The images in our TSI-CN dataset are collected from high quality and high-resolution real road scenes. The resolution of TSI-CN image is 2160$\times$3840, which is larger than that of other datasets.

\begin{figure*}[t]
	\centering
	\includegraphics[width=0.9\linewidth]{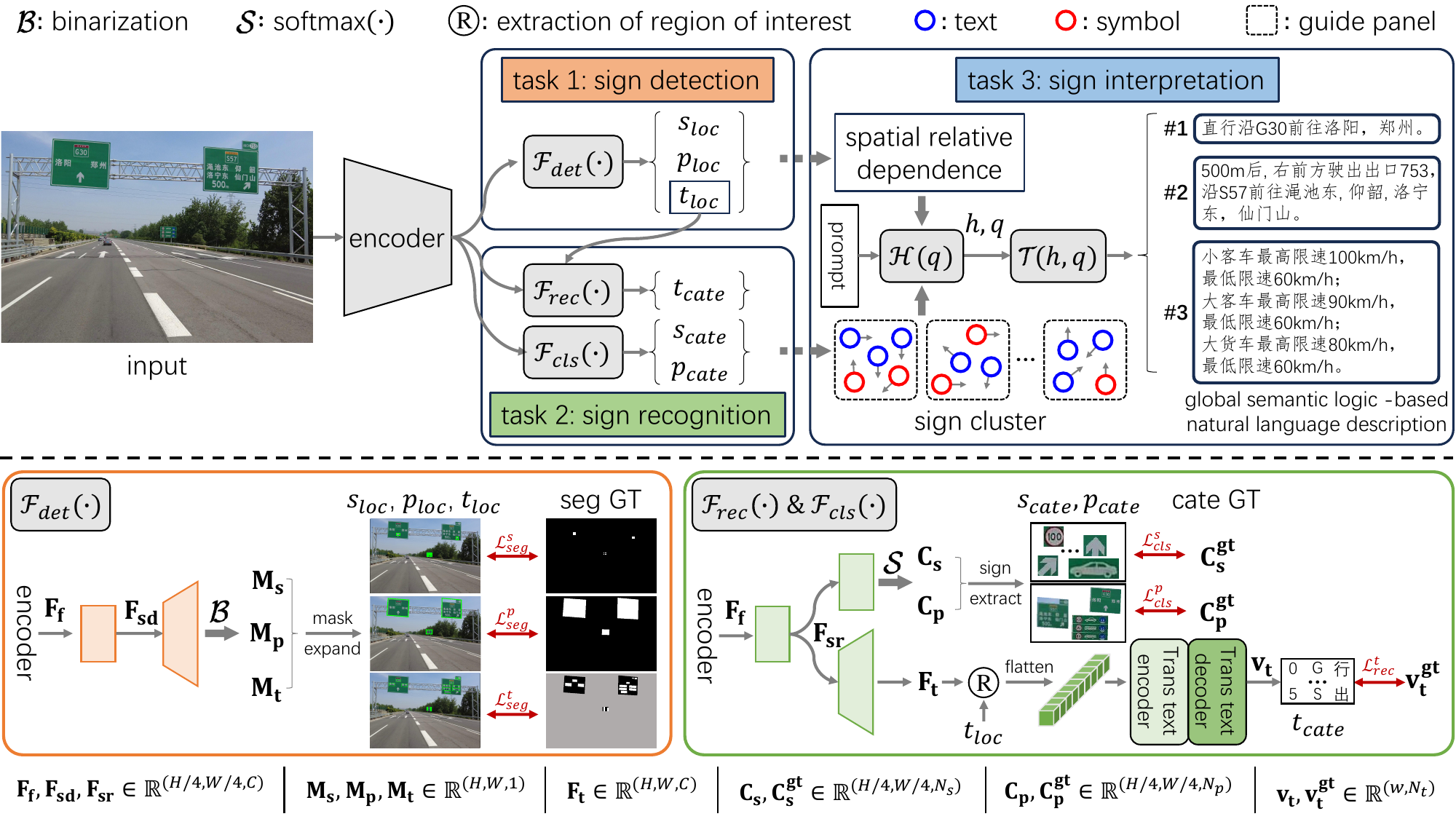}
	\vspace{-.1in}
	\caption{The overall architecture of TSI-arch. It consists of sign detection, recognition, and interpretation modules. The sign detection and recognition modules are illustrated in the bottom right and bottom left in detail and the interpretation module is described in Figure~\ref{fig:V4}.}
	\label{fig:V3}
	\vspace{-.1in}
\end{figure*}

\subsection{Data Split and Evaluation Metric}
\textbf{Data split.} TSI-CN Dataset is randomly split into two parts, namely training and test sets, which respectively contain 2,016 and 666 images. To ensure that the statistics (such as sign categories, the average value of sign resolutions) of the subset are almost the same. 


\textbf{Evaluation metrics.} We use Precision, Recall, and F-measure for TSI-arch's detection and recognition performance evaluation. As we all know, Precision, Recall, and F-measure are computed by the True Positive (TP), False Positive (FP), and False Negative (FN) samples. Here, for the detection task, the TP, FP, and FN are counted based on the sign boxes only. In the recognition task, the accuracy of the sign category is considered in the counting process. 

For the interpretation task, ROUGE~\cite{lin-2004-rouge} and Bleu~\cite{papineni2002bleu} metrics are adopted to evaluate the performance of recall and precision, respectively. Considering the above metrics cannot evaluate the semantic quality of the descriptions (such as both ``Go straight along G70 to Xi'an, Xianyang'' and ``Go straight along G70 to Xianyang, Xi'an'' are correct descriptions, but the ROUGE and Bleu measure them with different scores), we design the Soft Accuracy metric (SA) to evaluate the semantic quality by assessing whether the syntactic categories of the descriptions are consistent. It will measure the above example with the same score. For instance, since ``Go straight along G70 to Xi'an, Xianyang'' and ``Go straight along G70 to Xianyang, Xi'an'' belong to the syntax category ``Go straight--along--to'', SA scores them as 1, else 0.

\section{TSI Architecture}
\label{TSI Architecture}
To achieve the TSI task, we constructed a multi-task learning architecture (namely TSI-arch). In this section, the structure of TSI-arch, training and inference process, and optimization function will be described in detail.

\subsection{Overall Pipeline}
\label{Overall Pipeline}
As shown in Figure.~\ref{fig:V3}, TSI-arch consists of an encoder, sign detection module, sign recognition module, and sign interpretation module. The encoder is constructed based on the combination of backbone and feature pyramid network (FPN). It is responsible for the extraction of fused vision feature map $\mathbf{F}_f\in\mathbb{R}^{H/4, W/4, C}$ with multi-scaled context information, where $H$ and $W$ are the height and width of the input image and $C$ is the channel number of the feature map. The $\mathcal{F}_{det}(\cdot)$ and $\mathcal{F}_{cls}(\cdot)$ in sign detection and recognition modules take $\mathbf{F}_f$ as input to predict the sign locations $\{s_{loc},p_{loc},t_{loc}\}$ and the categories of symbols and guide panels $\{s_{cate},p_{cate}\}$ in parallel. For the text recognition, $\mathcal{F}_{rec}(\cdot)$ in the sign recognition module recognizes every single character $t_{cate}$ (including Arabic numeral, English, Chinese, and the other special characters) according to fused feature map $\mathbf{F}_f$ and text locations $t_{loc}$ gradually. With the sign detection and recognition results, the sign interpretation module first determines the spatial relative dependence of signs and groups symbols and texts that belong to the same sign together to generate chaotic sign clusters. It then organizes the internal semantic logic $h$ of signs according to the spatial relative dependence and chaotic sign clusters via $\mathcal{H}(q)$. In the end, the module feeds $h$ and the detected and recognized signs $q$ into $\mathcal{T}(h,q)$ for generating natural language descriptions with integrity traffic instruction information. The sign detection, recognition, and interpretation modules will be described in detail following.

\subsection{Sign Detection Module}
\label{Sign Detection Module}
\textbf{Module structure.} The structure can be found in the bottom left corner of Figure.~\ref{fig:V3}, which is designed for locating signs from road scene images. The function $\mathcal{F}_{det}(\cdot)$ in this module represents the detection mapping to be learned. It consists of a smooth convolutional layer and a segmentation head. The former is used for smoothing the gap between high-level and low-level features via the stack of a 3$\times$3 filer, BN~\cite{ioffe2015batch}, and ReLU~\cite{glorot2011deep} to generate smooth feature map $\mathbf{F}_{sd}\in\mathbb{R}^{H/4, W/4, C}$. The head takes $\mathbf{F}_{sd}$ to segment shrink-masks ($\mathbf{M}_s, \mathbf{M}_p,$ and $\mathbf{M}_t\in\mathbb{R}^{H, W, 1}$) of signs through two transposed convolutions with 2$\times$2 filer and binarization operator $\mathcal{B}$. The final bounding boxes of signs are obtained by expanding shrink-masks to original regions.

\textbf{Training stage.} the label of shrink-mask is generated by shrinking the sign bounding box using the polar minimum distance algorithm~\cite{yang2022cm}. The shrinking offset $d_s$ is computed from the coordinates of the sign center point and a series of dense original sign box points:
\begin{eqnarray}
d_s=\frac{1}{2}\mathrm{min}(\left \| p_{cp},p_n \right \|_{2}^{2} ),n=1,2,...,N,
\end{eqnarray} 
where $p_{cp}$ and $p_n$ are the coordinates of the sign center point and sign dense box points. $N$ is the number of dense box points. Based on sign shrink-masks, the optimization function of detection module can be formulated as follows:
\begin{eqnarray}
\mathcal{L}_{det}=\mathcal{L}^{s}_{seg}+\mathcal{L}^{p}_{seg}+\mathcal{L}^{t}_{seg},
\label{Eqn:E1}
\end{eqnarray} 
where Dice loss~\cite{milletari2016v} is adopted as $\mathcal{L}_{seg}$ in Eqn.~\ref{Eqn:E1}.

\textbf{Inference stage.} The sign box can be obtained easily by expanding the predicted shrink-mask contour using the expanding distance $d_e$:
\begin{eqnarray}
d_e=\mathrm{min}(\left \| p_{cp},p_m \right \|_{2}^{2} ),m=1,2,...,M,
\end{eqnarray} 
where $p_{cp}$ and $p_m$ are the coordinates of the shrink-mask center point and shrink-mask dense contour points. $M$ is the number of dense contour points. 

\subsection{Sign Recognition Module}
\label{Sign Recognition}    
\textbf{Module structure.} The structure can be found in the bottom right corner of Figure.~\ref{fig:V3}, which is responsible for classifying the categories of symbols and guide panels and recognizing texts via $\mathcal{F}_{cls}(\cdot)$ and $\mathcal{F}_{rec}(\cdot)$, respectively. The same as $\mathcal{F}_{det}(\cdot)$, the function $\mathcal{F}_{cls}(\cdot)$ smooths multi-scaled features via a smooth layer and obtains the feature map $\mathbf{F}_{sr}\in\mathbb{R}^{H/4, W/4, C}$. It then predicts category probability maps of symbols $\mathbf{C}_{s}\in\mathbb{R}^{H/4, W/4, N_s}$ and panels $\mathbf{C}_{p}\in\mathbb{R}^{H/4, W/4, N_p}$ via two convolutions with 3$\times$3 filer and softmax operator $\mathcal{S}$. Notably, the category classification and the sign detection are conducted in parallel, which helps simplifies the framework's complexity. As for $\mathcal{F}_{rec}(\cdot)$, it is a function mapping for predicting character distribution probabilities. The function takes the flatten crop features that extracted from $\mathbf{F}_{sr}$ according to text location information $t_{loc}$ as input. It focuses on every single character via transformer encoder and decoding a series of character distribution probability vector $\mathbf{v}_t\in\mathbb{R}^{w,N_t}$ through transformer decoder, where the encoder and decoder are consists of a multi head attention layer~\cite{vaswani2017attention} with 8 heads. The final recognized text can be transformed from $\mathbf{v}_t$ through the vec2word algorithm in~\cite{mikolov2013efficient}.

\textbf{Training stage.} The label of the symbol $\mathbf{C}_{s}^{gt}\in\mathbb{R}^{H/4, W/4, N_s}$ and panel $\mathbf{C}_{p}^{gt}\in\mathbb{R}^{H/4, W/4, N_p}$ categories are composed of $\frac{H}{4}\times \frac{W}{4} $ one-hot classification vector, where $N_s$ and $N_s$ are the category number of symbol and panel, respectively. For the generation of text label, we use word2vec algorithm proposed in~\cite{mikolov2013efficient} to embedding character into vector representation $\mathbf{v}_{t}^{gt}\in\mathbb{R}^{w,N_t}$, where $w$ and $N_t$ are the length of the flatten features and vector dimension. For the recognition task of symbol, text, and panel, we use cross-entropy loss to formulate the optimization function:
\begin{eqnarray}
\mathcal{L}_{rec}=\mathcal{L}^{s}_{ce}+\mathcal{L}^{p}_{ce}+\mathcal{L}^{t}_{ce}.
\label{Eqn:E2}
\end{eqnarray}

\textbf{Inference stage.} The symbol and panel category vectors are sampled from $\mathbf{C}_{s}$ and $\mathbf{C}_{p}$ by the symbol and panel box center positions at first. Then, category vectors are transformed into one-hot vectors via the softmax function and mapped to the corresponding categories with a category dict. The same as the symbol and panel's category prediction, a one-hot character vector is generated by the softmax function according to the outputted character distribution probability vector $\mathbf{v}_t$, and the character can be retrieved from a character dict.
\begin{figure}[t]
	\centering
	\includegraphics[width=0.99\linewidth]{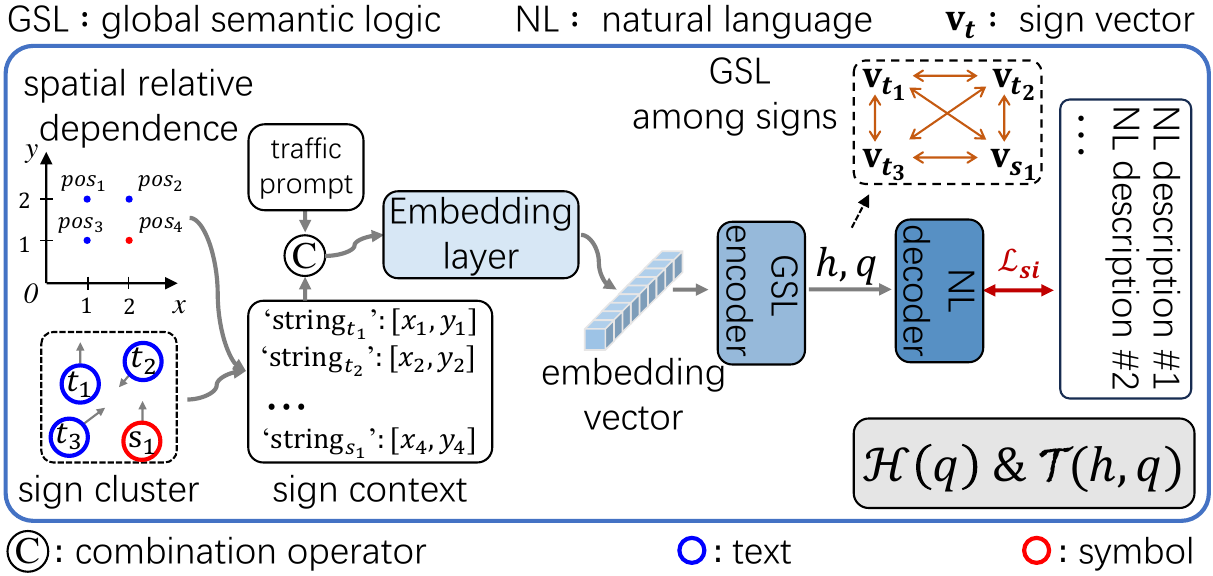}
	\vspace{-.1in}
	\caption{The illustration of sign interpretation module structure, where sign cluster consists of recognized sign results.}
	\label{fig:V4}
	\vspace{-.1in}
\end{figure}
\subsection{Sign Interpretation}
\label{Sign Interpretation}
\textbf{Module structure and inference stage.} The structure can be found in Figure.~\ref{fig:V4}, which is used for organizing the global semantic logic between detection and recognition signs $q$. This module combines spatial relative dependence $\{s_{loc},t_{loc},p_{loc}\}$ and sign cluster $\{s_{cate},t_{cate},p_{cate}\}$ for generating sign context first. Here, the spatial relative dependence is obtained from sign locations, and the distances between signs are normalized as 1. The sign cluster consists of recognition strings of signs (including symbols and texts) within the same guide panel. Then, the traffic prompt and sign context are embedded into vectors via the embedding layer that consists of a standard embedding operator implemented by Pytorch and an LSTM~\cite{graves2012long} layer. Next, $\mathcal{H}(\cdot)$, a stacked GLM block~\cite{DBLP:conf/acl/DuQLDQY022}, takes the embedding vectors and sign context as input to analyze the global semantic logic among signs and to organize the logic as a hidden state $h$. In the end, $\mathcal{T}(\cdot)$ decodes the hidden state $h$ to a series of word vectors through the combination of GLM blocks and a fully connected layer, and the final natural language description is generated by the vec2word algorithm.
\begin{table*}[h]
	\renewcommand{\arraystretch}{1}
	\setlength{\tabcolsep}{2mm}
	\centering
	\footnotesize 
	\caption{The detection performance of the models for dealing with traffic signs of different scales. `60' and `80' denote the input image is resized to 0.6 and 0.8 times the original size in the training process. `S', 'M', and 'L' mean scaling the input image to 736, 1024, and 1280 along the short side in the inference stage.}
	\vspace{-.1in}
	\begin{tabular}{l|ccc|ccc|ccc}
		\Xhline{1.5pt}
		\multirow{2}{*}{Methods} & \multicolumn{3}{c|}{symbol}                                              & \multicolumn{3}{c|}{text}                                                & \multicolumn{3}{c}{panel}                                         \\ \cline{2-10} 
		& \multicolumn{1}{c}{Precision} & \multicolumn{1}{c}{Recall} & F-measure & \multicolumn{1}{c}{Precision} & \multicolumn{1}{c}{Recall} & F-measure & \multicolumn{1}{c}{Precision} & \multicolumn{1}{c}{Recall} & F-measure \\ 
		\hline
		$\mathrm{TSI}_{40-S}$   & \multicolumn{1}{c}{60.82}    & \multicolumn{1}{c}{45.62} & 52.13    & \multicolumn{1}{c}{66.10}    & \multicolumn{1}{c}{53.01} & 58.83    & \multicolumn{1}{c}{\cellcolor{gray!30}89.33}    & \multicolumn{1}{c}{\cellcolor{gray!30}90.28} &\cellcolor{gray!30} {\textcolor{magenta}{90.07}}    \\ 
		$\mathrm{TSI}_{40-M}$   & \multicolumn{1}{c}{76.04}    & \multicolumn{1}{c}{63.55} & 69.23    & \multicolumn{1}{c}{75.02}    & \multicolumn{1}{c}{66.66} & 70.59    & \multicolumn{1}{c}{82.73}    & \multicolumn{1}{c}{93.43} & 87.76    \\ 
		$\mathrm{TSI}_{40-L}$   & \multicolumn{1}{c}{76.60}    & \multicolumn{1}{c}{67.83} & 71.95    & \multicolumn{1}{c}{76.18}    & \multicolumn{1}{c}{72.37} & 74.23    & \multicolumn{1}{c}{78.61}    & \multicolumn{1}{c}{84.68} & 85.09    \\ \hline
		$\mathrm{TSI}_{60-S}$   & \multicolumn{1}{c}{63.97}    & \multicolumn{1}{c}{48.11} & 54.92    & \multicolumn{1}{c}{67.44}    & \multicolumn{1}{c}{54.22} & 60.11    & \multicolumn{1}{c}{88.70}    & \multicolumn{1}{c}{89.26} & 88.98    \\ 
		$\mathrm{TSI}_{60-M}$   & \multicolumn{1}{c}{74.58}    & \multicolumn{1}{c}{65.74} & 69.88    & \multicolumn{1}{c}{76.27}    & \multicolumn{1}{c}{69.05} & 72.48    & \multicolumn{1}{c}{82.26}    & \multicolumn{1}{c}{91.87} & 86.80    \\ 
		$\mathrm{TSI}_{60-L}$   & \multicolumn{1}{c}{\cellcolor{gray!30}80.29}    & \multicolumn{1}{c}{\cellcolor{gray!30}77.09} & {\cellcolor{gray!30}\textcolor{magenta}{78.66}}    & \multicolumn{1}{c}{\cellcolor{gray!30}76.46}    & \multicolumn{1}{c}{\cellcolor{gray!30}76.90} & {\cellcolor{gray!30}\textcolor{magenta}{76.68}}    & \multicolumn{1}{c}{74.57}    & \multicolumn{1}{c}{92.91} & 85.14    \\ \hline
		$\mathrm{TSI}_{80-S}$   & \multicolumn{1}{c}{66.37}    & \multicolumn{1}{c}{43.82} & 52.79    & \multicolumn{1}{c}{66.11}    & \multicolumn{1}{c}{50.89} & 57.51    & \multicolumn{1}{c}{90.08}    & \multicolumn{1}{c}{85.19} & 87.57    \\ 
		$\mathrm{TSI}_{80-M}$   & \multicolumn{1}{c}{75.69}    & \multicolumn{1}{c}{60.16} & 67.04    & \multicolumn{1}{c}{73.69}    & \multicolumn{1}{c}{65.31} & 69.25    & \multicolumn{1}{c}{86.75}    & \multicolumn{1}{c}{91.45} & 89.04    \\ 
		$\mathrm{TSI}_{80-L}$   & \multicolumn{1}{c}{80.11}    & \multicolumn{1}{c}{71.81} & 75.74    & \multicolumn{1}{c}{74.59}    & \multicolumn{1}{c}{71.57} & 73.05      & \multicolumn{1}{c}{81.77}    & \multicolumn{1}{c}{92.60} & 86.85    \\ 
		\Xhline{1.5pt}
	\end{tabular}
	\vspace{-.1in}
	\label{tab:T2}
\end{table*}
\begin{table*}[h]
	\renewcommand{\arraystretch}{1}
	\setlength{\tabcolsep}{1mm}
	\centering
	\footnotesize
	\caption{The recognition performance of the model ($\mathrm{TSI}_{60-L}$).  Considering there are lots of different symbols in the TSI-CN dataset, some representative symbols are picked up to show the model performance (e.g., i78, i54, etc.) conveniently.  `OA' denotes the overall accuracy of all kinds of components.  `N', `C', and `E' are Arabic numerals, Chinese, and English characters.}
	\vspace{-.1in}
	\begin{tabular}{c|c|c|c|c|c|c|c|c|c|c|c|c|c}
		\Xhline{1.5pt}
		\multirow{2}{*}{\begin{tabular}[c]{@{}c@{}}{\textbf{symbol}}\\ \textbf{Acc}\end{tabular}} & i78    & i54    & i81    & i79    & a1     & a3     & a5     & a7     & p4                                                                  & p34    & p39    & p36    & OA    \\ \cline{2-14} 
		& 29.99 & 48.00   & 67.92 & 74.99 & 66.18 & 71.42 & 79.41 & 76.56 & 79.99                                                              & 78.04 & 59.99 & 44.44 & 71.82 \\ \hline
		\multirow{2}{*}{\begin{tabular}[c]{@{}c@{}}\textbf{panel}\\ \textbf{Acc}\end{tabular}}  & 0      & 1      & 2      & 3      & 4      & 5      & 6      & OA    & \multirow{2}{*}{\begin{tabular}[c]{@{}c@{}}\textbf{text}\\ \textbf{Acc}\end{tabular}} & N      & C      & E      & OA    \\ \cline{2-9} \cline{11-14} 
		& 26.67 & 86.46 & 80.24 & 88.35 & 93.75 & 84.35 & 30.76 & 84.75 &                                                                     &    60.19    &    62.15    &    60.90    & 62.93 \\
		\Xhline{1.5pt}
	\end{tabular}
	\vspace{-.1in}
	\label{tab:T3}
\end{table*}
\begin{figure*}[!h]
\centering
\begin{minipage}[t]{0.999\textwidth}
	\centering
	\begin{subfigure}[t]{0.49\textwidth}
		\centering
		\includegraphics[scale=0.29]{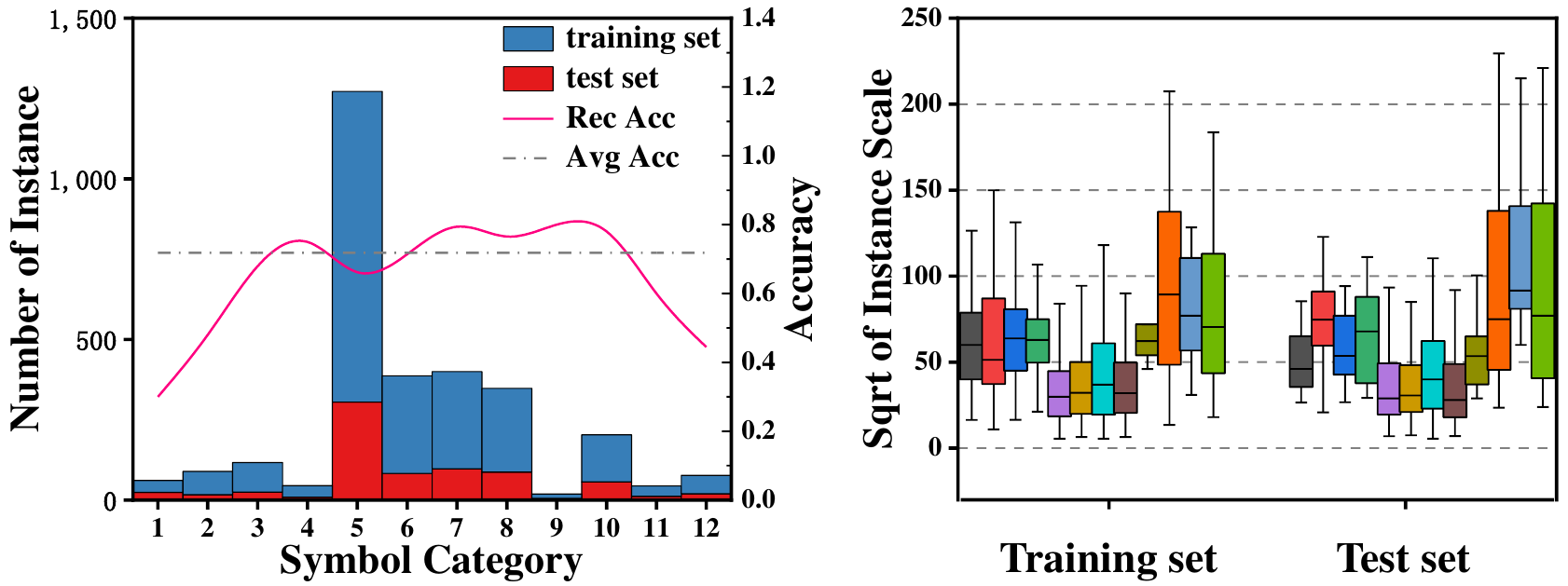}
		\caption{symbol recognition performance}
		\label{fig:V5-1}
	\end{subfigure}
	\begin{subfigure}[t]{0.49\textwidth}
		\centering
		\includegraphics[scale=0.29]{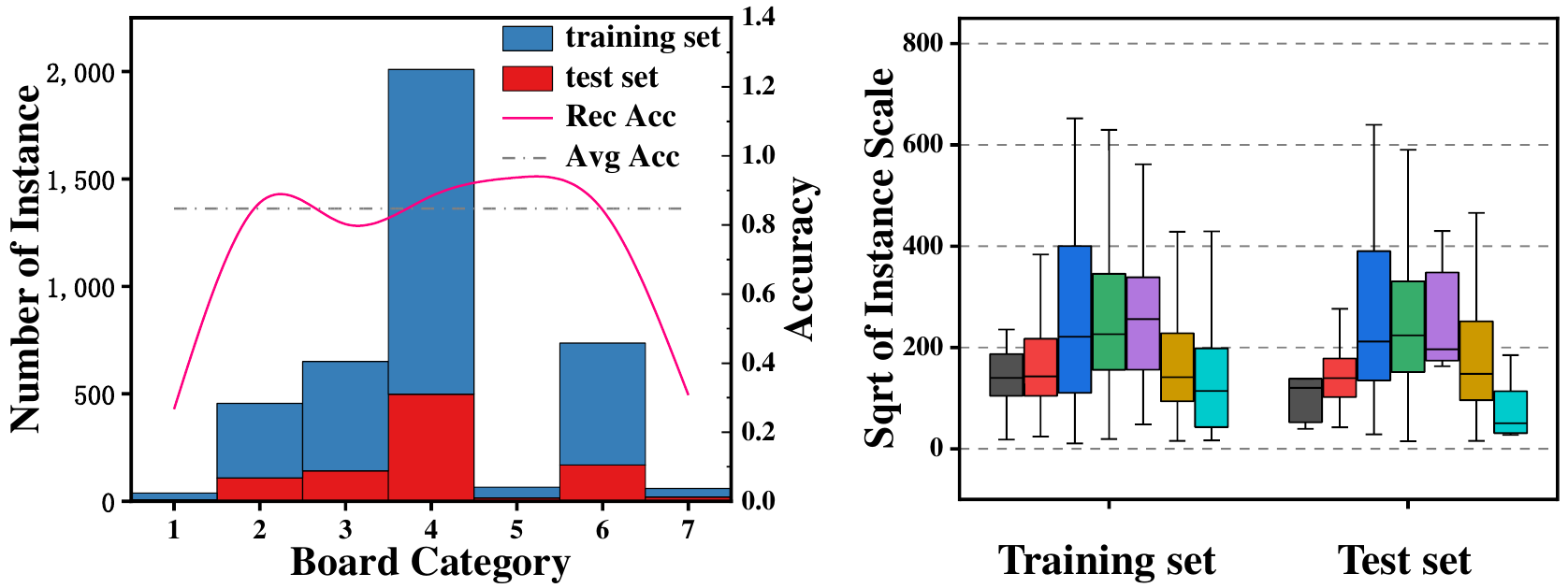}
		\caption{guide panel recognition performance}
		\label{fig:V5-2}
	\end{subfigure}
\end{minipage}
\vspace{-.1in}
\caption{The illustration of sign recognition performance and the distribution of instance number, and the distribution of instance scale.}
\label{fig:V5}
\vspace{-.1in}
\end{figure*}

\textbf{Training stage.} The data label is a natural language sentence that is organized based on signs within a guide panel according to the Chinese design criteria of road traffic signs. The optimization function $\mathcal{L}_{int}$ is formulated by cross-entropy loss just like the recognition loss function.


\begin{figure*}[t]
	\centering
	\includegraphics[width=0.99\linewidth]{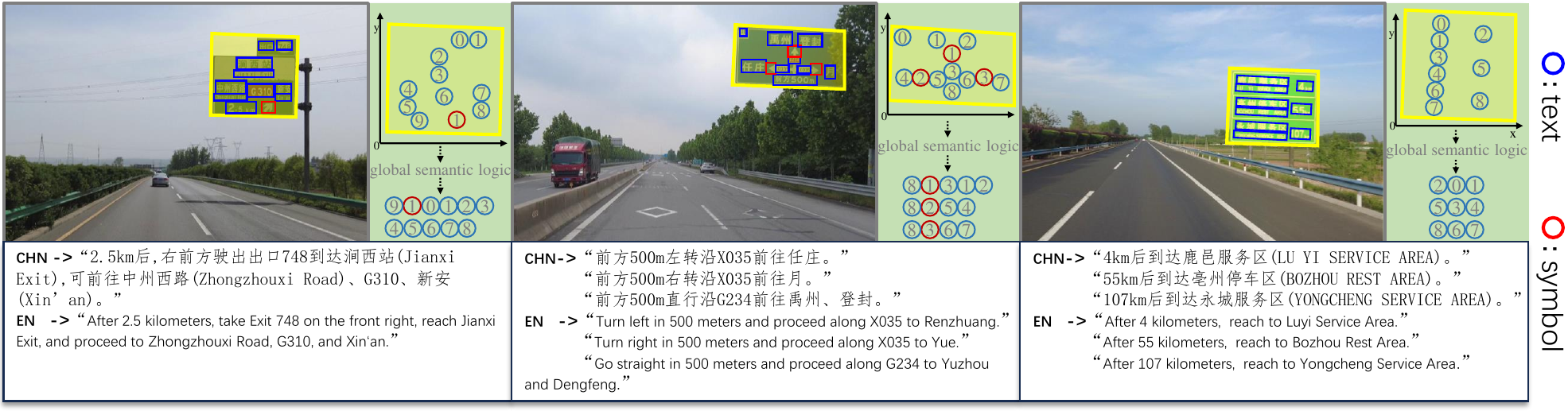}
	\vspace{-.1in}
	\caption{The visualization of the interpretation process and the global semantic logic-based natural language description.}
	\label{fig:V6}
	\vspace{-.15in}
\end{figure*}
\section{Experiments}
As described in Section~\ref{TSI Architecture}, the designed multi-task framework involves the three aspects of the detection, recognition, and interpretation of traffic signs. In this section, we verify the effectiveness of the framework on the TSI-CN dataset via the evaluation of all three aspects.

\subsection{Implementation Details}
In the experiments, the combination of ResNet-18~\cite{he2016deep} and FPN is adopted as the encoder of TSI-arch, where the channel number of outputted fused vision feature map $\mathbf{F}_f$ is set to 64. In the data pre-processing stage, the training samples are increased by the following augmentation strategies: (1) random scaling (including image size and aspect ratio); (2) random rotating in the range of (-10°, 10°); (3) random cropping and padding. To optimize the proposed network, the Adam algorithm~\cite{kingma2014adam} is deployed to optimize the model. For learning rate, it is initialized as 0.001 and adjusted through the `polylr' strategy~\cite{yu2018bisenet}. All experimental results are obtained through training TSI-arch by 200 epochs and 24 batch sizes.

\subsection{Results Analysis on Sign Detection}
Here, we show the detection performance of TSI-arch on signs in Table~\ref{tab:T2}. As mentioned in Section~\ref{TSI-CN Dataset}, the TSI-CN dataset consists of large-scale images, which is beneficial to the effective detection of some small traffic signs (such as distant symbols and texts). However, the guide panel always enjoys a larger scale than the symbol and text (it even can be up to a quarter of the input image size), which makes it hard for TSI-arch to locate panels integrally and accurately. Therefore, the overall performance of the models for dealing with different scaled images are listed to explore the influence brought by image and instance scales. 

Concretely, the input is resized to 0.4, 0.6, and 0.8 times the original size in the training process. For the inference stage, we scale the image to 736 ($S$), 1024 ($M$), and 1280 ($L$) along the short side. We observe that the detection performance of the model on symbols and texts gains improvements continually by tuning the image size larger in the inference process and keeping it fixed in the training stage, while it shows an opposite trend on the detection performance of the guide panel. It can be found that $\mathrm{TSI}_{60-L}$ achieves the best results on the detection of symbols and texts (78.66\% and 76.68\% in F-measure) and $\mathrm{TSI}_{40-S}$ achieves the best results on guide panel (90.05\% in F-measure). The above experiments verify the proposed TIS-arch can detect different signs effectively and it is important for exploring the influence brought by image scales to different signs. 

\subsection{Results Analysis on Sign Recognition}
Table~\ref{tab:T3} lists the recognition results of our model on signs. To show the performance of symbols conveniently, some representative of them (e.g., i78, i54, etc.) are picked out in this experiment. It is found that TSI-arch achieves superior performance of 71.82\% in average accuracy on symbols, while the model performs badly on some of them (such as i78, i54, and p36). For the guide panel, a similar conclusion can be observed. To explore the reason behind this phenomenon, we analyze the relationship between the recognition performance and the distributions of instance number and scale in Fig~\ref{fig:V5}. It can be observed that large instance numbers can bring performance gains. Meanwhile, a big instance scale also can ensure our model achieves a competitive recognition accuracy on symbols even though there are few training samples in the TSI-CN dataset. For text recognition, we list the accuracy of different kinds of characters in Table~\ref{tab:T3}, since the complex stroke and categories, the recognition accuracy in Chinese is behind the numbers and English characters. 
\begin{table}[]
	\renewcommand{\arraystretch}{1.1}
	\setlength{\tabcolsep}{1.75mm}
	\centering
	\footnotesize
	\caption{Performance of the models with different settings. `R-1', `R-2', `R-$l$', `B-4', and `SA' are Rouge-1, Rouge-2, Rouge-$l$, Bleu-4, and soft accuracy, respectively. `SRD' and `TP' are spatial relative dependence and traffic prompt. }
	\vspace{-.1in}
	\begin{tabular}{c|l|ccccc}
		\Xhline{1.5pt} 
		& Methods                      & R-1 & R-2 & R-$l$ & B-4 & SA \\ \hline
		\multirow{3}{*}{\rotatebox[]{90}{{detection}}}    & Baseline                     &  {69.52}          & {46.86}          & {59.21}          & {44.28}          & 44.70      \\ \
		& Baseline + TP       &  {69.32}          & {46.46}          & {59.58}          & {44.84}          & 47.03       \\ \
		& Baseline + TP + SRD &   {\cellcolor{gray!30}\textcolor{magenta}{70.69}} & {\cellcolor{gray!30}\textcolor{magenta}{47.58}} & {\cellcolor{gray!30}\textcolor{magenta}{61.22}} & \cellcolor{gray!30}\textcolor{magenta}{45.62}          & \cellcolor{gray!30}\textcolor{magenta}{48.31} \\ \hline
		\multirow{1}{*}{\rotatebox[]{90}{{GT}}} 
		& Baseline + TP + SRD  &  {92.08}          & {82.11}          & {{88.30}} & {84.48}          & 76.06  \\ 
		\Xhline{1.5pt}
	\end{tabular}
	\vspace{-.1in}
	\label{tab:T4}
\end{table}
\begin{table}[]
	\renewcommand{\arraystretch}{1}
	\setlength{\tabcolsep}{.8mm}
	\centering
	\footnotesize
	\caption{Performance comparison with related methods on different sub-tasks. `F' and `Acc' are F-measure and accuracy metrics.}
	\vspace{-.1in}
	\begin{tabular}{l|cc|cc|cc|ccccc}  
		\Xhline{1.5pt}
		\multirow{2}{*}{Methods} & \multicolumn{2}{c|}{symbol}          & \multicolumn{2}{c|}{text}            & \multicolumn{2}{c|}{panel}           & \multicolumn{5}{c}{interpretation}                                  \\ \cline{2-12} 
		& \multicolumn{1}{c}{F} & Acc & \multicolumn{1}{c}{F} & Acc & \multicolumn{1}{c}{F} & Acc & \multicolumn{1}{c}{R-1} & \multicolumn{1}{c}{R-2} & R-$l$  & B-4 & SA\\ \hline
		
		Faster-RCNN~\cite{ren2015faster}& \multicolumn{1}{c}{49}          &  44   & \multicolumn{1}{c}{75}          &  --   & \multicolumn{1}{c}{60}          &  59   & \multicolumn{1}{c}{--}        & 
		\multicolumn{1}{c}{--}        &    --     & -- & -- \\ 
		
		YOLO~\cite{redmon2016you}& \multicolumn{1}{c}{46}          &  40   & \multicolumn{1}{c}{{\cellcolor{gray!30}\textcolor{magenta}{78}}}          &  --   & \multicolumn{1}{c}{47}          &  46   & \multicolumn{1}{c}{--}        & \multicolumn{1}{c}{--}        &    --    & -- & --  \\ 
		
		LeafText~\cite{yang2023text}                 & \multicolumn{1}{c}{--}          &  --   & \multicolumn{1}{c}{68}          &  46   & \multicolumn{1}{c}{--}          &  --   & \multicolumn{1}{c}{--}        & \multicolumn{1}{c}{--}        &     --     & -- & --\\ \hline
		
		TSI (\textbf{Ours})               & \multicolumn{1}{c}{\cellcolor{gray!30}\textcolor{magenta}{79}}          &  \cellcolor{gray!30}\textcolor{magenta}{72}   & \multicolumn{1}{c}{77}          & \cellcolor{gray!30}\textcolor{magenta}{63}     & \multicolumn{1}{c}{\cellcolor{gray!30}\textcolor{magenta}{85}}          &  \cellcolor{gray!30}\textcolor{magenta}{85}   & \multicolumn{1}{c}{\cellcolor{gray!30}\textcolor{magenta}{71}}        & \multicolumn{1}{c}{\cellcolor{gray!30}\textcolor{magenta}{48}}        &  \cellcolor{gray!30}\textcolor{magenta}{61}     & \cellcolor{gray!30}\textcolor{magenta}{46} &  \cellcolor{gray!30}\textcolor{magenta}{48}  \\
		\Xhline{1.5pt}
	\end{tabular}
	\vspace{-.1in}
	\label{tab:T5}
\end{table}
\subsection{Results Analysis on Sign Interpretation}
\textbf{Quantitative results.} We explore the performance of the interpretation modules with different settings in Table~\ref{tab:T4}, where `detection' and `gt' mean the results are obtained based on the predicted and ground-truth signs, respectively. For the `detection' results, it is observed that the combination of `TP' and `SRD' brings improvement in all metrics, which proves that it can encourage the TSI-arch to organize the global semantic logic among signs and interpret them as natural language descriptions more accurately. Meanwhile, our model gains 2.3\% improvement with the help of `TP' compared with the baseline. For the `gt' results, the proposed TSI-arch achieves high-quality interpretation descriptions, which verifies the interpretation module's effectiveness without influences from detection and recognition.

\textbf{Visualization results.} Figure~\ref{fig:V6} shows the interpretation process and the final results. It can be observed that the TSI-arch can detect and recognize traffic-related symbols, texts, and guide panels accurately from real road scenes even though those sign scales vary dramatically, which provides support for the analysis of global semantic logic and the generation of natural language descriptions.

\subsection{Comparison with Related Methods}
Considering TSI is a novel task, to show the superior performance of the TSI-arch, we show the performance comparison with traditional object detection methods (e.g., Faster-RCNN~\cite{ren2015faster} and YOLO~\cite{redmon2016you}) on the symbol detection and with the OCR method (LeafText~\cite{yang2023text}) on text recognition, respectively. It can be observed in Table~\ref{tab:T5}, benefiting from the advantages of multi-task optimization, TSI-arch achieves outperforms detection task-only (Faster-RCNN~\cite{ren2015faster} and YOLO~\cite{redmon2016you}) and text-related only methods in both detection and recognition accuracy.

\section{Conclusion}
In this paper, we have presented a novel task, namely traffic sign interpretation (TSI), to interpret accurate traffic instruction information via natural language like a human from real road scenes, which can promote the progress of the traffic sign recognition community while providing support for the development of autonomous and assistant driving systems. Meanwhile, we explore and design a multi-task framework to detect and recognize traffic signs, analyze the internal semantic logic among them, and interpret them as a natural language. Furthermore, a TSI-CN dataset is constructed to fulfill the research and evaluation of the TSI framework, which encourages more researchers to participate in the research of the TSI task. Experimental results demonstrate the TSI task is achievable and the framework has achieved superior performance in all aspects of traffic sign detection, recognition, and interpretation. In future work, we will concentrate on the practical application environment and requirement-related challenges (e.g. dark light, motion blur, and key information retrieval) in the research of TSI.

{
    \small
    \bibliographystyle{ieeenat_fullname}
    \bibliography{main}
}


\end{document}